\documentclass{article}
\usepackage{ijcai25}

\usepackage{times}
\usepackage{soul}
\usepackage{url}
\usepackage[hidelinks]{hyperref}
\usepackage[utf8]{inputenc}
\usepackage[small]{caption}
\usepackage{graphicx}
\usepackage{amsmath}
\usepackage{amsthm}
\usepackage{booktabs}
\usepackage{algorithm}
\usepackage{algorithmic}
\usepackage[switch]{lineno}
\usepackage{subcaption}  

\title{Dual-Agent Reinforcement Learning for Automated Feature Generation}

\author{
Wanfu Gao$^{1,2}$\and
Zengyao Man$^{1,2}$\and
Hanlin Pan$^{1,2}$\thanks{Corresponding author}\And
Kunpeng Liu$^{3}$
\affiliations
$^1$College of Computer Science and Technology, Jilin University, China\\
$^2$Key Laboratory of Symbolic Computation and Knowledge Engineering of Ministry of Education, Jilin University, China\\
$^3$Department of Computer Science, Portland State University, Portland, OR 97201 USA\\
\emails
gaowf@jlu.edu.cn,
manzy23@mails.jlu.edu.cn,
panhl23@mails.jlu.edu.cn,
kunpeng@pdx.edu
}

\begin{document}

\maketitle

\begin{abstract}
Feature generation involves creating new features from raw data to capture complex relationships among the original features, improving model robustness and machine learning performance. Current methods using reinforcement learning for feature generation have made feature exploration more flexible and efficient. However, several challenges remain: first, during feature expansion, a large number of redundant features are generated. When removing them, current methods only retain the best features each round, neglecting those that perform poorly initially but could improve later. Second, the state representation used by current methods fails to fully capture complex feature relationships. Third, there are significant differences between discrete and continuous features in tabular data, requiring different operations for each type. To address these challenges, we propose a novel dual-agent reinforcement learning method for feature generation. Two agents are designed: the first generates new features, and the second determines whether they should be preserved. A self-attention mechanism enhances state representation, and diverse operations distinguish interactions between discrete and continuous features. The experimental results on multiple datasets demonstrate that the proposed method is effective\footnote{The code is available at https://github.com/extess0/DARL}.
\end{abstract}

\section{Introduction}

\begin{figure}[t]
\centering
\includegraphics[width=0.95\columnwidth]{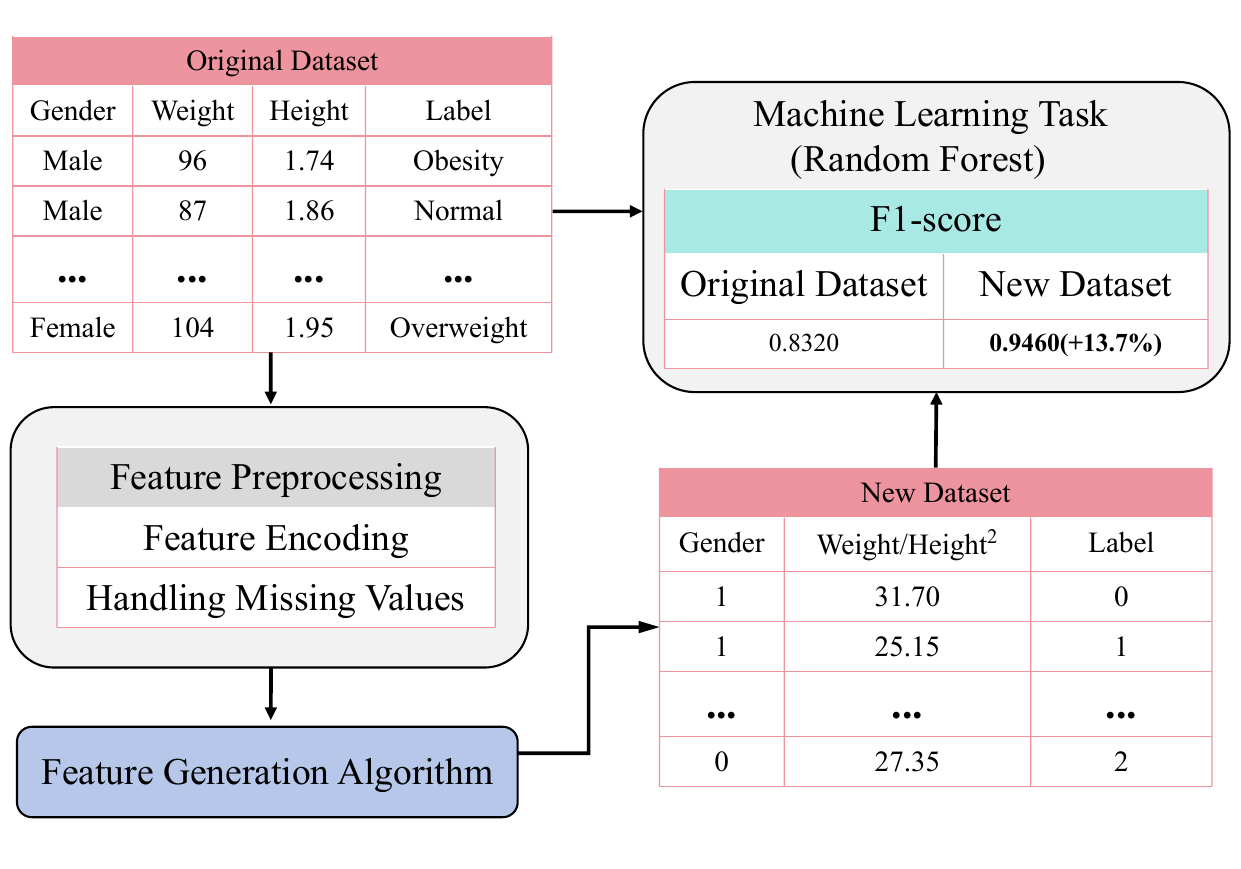}
\caption{After data preprocessing, the dataset is transformed using a feature generation algorithm, resulting in a new dataset that significantly improved the \textit{F1-score} in downstream machine learning tasks compared to the original dataset.}
\label{fig_1}
\end{figure}

Feature generation in machine learning is the process of combining original data features to create new ones. These new features can capture complex relationships between original features, enhance model robustness, improve data representation, and ultimately improve the performance of machine learning tasks.
As Figure \ref{fig_1} shows, there are several features: gender, weight and height, and our goal is to predict whether a person is healthy based on these features. Feature generation methods can help create a new feature $\frac{\text{weight}}{\text{height}^2}$, known as the Body Mass Index (BMI) \cite{BMI}, which can lead to better predictive performance. However, manual feature generation by domain experts is labor-intensive and does not accurately capture the relationships among a large number of features. Therefore, automated methods for feature generation are naturally needed.
Some traditional feature generation methods, such as FCTree \cite{FCTree} and FICUS \cite{FICUS}, are still influenced by domain knowledge, lacking flexibility and adaptability. Current methods using reinforcement learning (RL) for feature generation have made the exploration of features more flexible and efficient \cite{khurana2018feature,wang2022group}. Despite these methods utilizing reinforcement learning for feature generation showing potential improvement, several challenges still exist. First, in the feature expansion phase of reinforcement learning, a large number of redundant features are generated that need to be removed. Current RL-based feature generation methods typically involve feature selection based on mutual information after generating features, selecting the top \textit{K} features with the most information for subsequent feature generation. However, this method only considers the best features in each round, neglecting features that initially perform poorly but contribute better performance in later rounds. Second, RL agents must take states as input and current state representation methods primarily focus on autoencoders and graph convolutional networks \cite{xiao2022self}, but these methods fail to sufficiently capture the complex relationships between features. Third, there are significant differences between discrete and continuous features, and it requires different operations for different types of features. For example, mathematical operations like logarithmic or exponential operations are only suitable for continuous features while cross-product is only suitable for discrete features. Most existing methods ignore interactions between discrete and continuous features.

To deal with these challenges, in this paper, we propose a Dual-Agent Reinforcement Learning (DARL) method for automated feature generation. \textbf{To deal with the first challenge,} the concept of hierarchical reinforcement learning \cite{DBLP,kulkarni2016hierarchical,EHRL} is used to decompose complex tasks into smaller, simpler sub-tasks. Specifically, we designed two agents for the reinforcement learning framework: the first agent is responsible for feature generation, and the second agent is responsible for feature discrimination. Feature discrimination is used to determine whether the generated features are worth retaining. By integrating the feature discrimination process into the reinforcement learning framework, an optimal subset of features is obtained that performs well in the current decision step and offers better long-term rewards, as reinforcement learning aims to maximize cumulative rewards. The method effectively explores the feature space to generate a superior set of features. Compared to traditional mutual information methods, this method achieves better results.
\textbf{To deal with the second challenge,} the representation of states in RL is enhanced using the Transformer model \cite{vaswani2017attention}. Through self-attention mechanisms, the state representation in RL can reveal correlations between each feature and provide rich information for RL algorithms to better understand complex feature relationships.
\textbf{As for the third challenge,} we use different arithmetic operations to capture three types of relationships, i.e., the relationship between discrete and discrete features, the relationship between continuous and continuous features, as well as the relationship between discrete and continuous features.

In summary, this paper presents a novel dual-agent reinforcement learning feature generation method. 
Through the dual-agent reinforcement learning process, features are generated with better performance and higher interpretability. Our main contributions include:

\begin{itemize}
\item The method Dual-Agent Reinforcement Learning (DARL) is used for feature generation, where the first agent is responsible for generating features, and the second agent is responsible for preserving useful features and removing redundant ones.
\item We propose to use the self-attention mechanism for a reinforcement learning state, which can lead to better embedding representations.
\item We propose to distinguish between discrete and continuous feature interactions, which enables the generation of more interpretable features.
\end{itemize}

\begin{figure*}[t]
\centering
\includegraphics[width=0.8\textwidth]{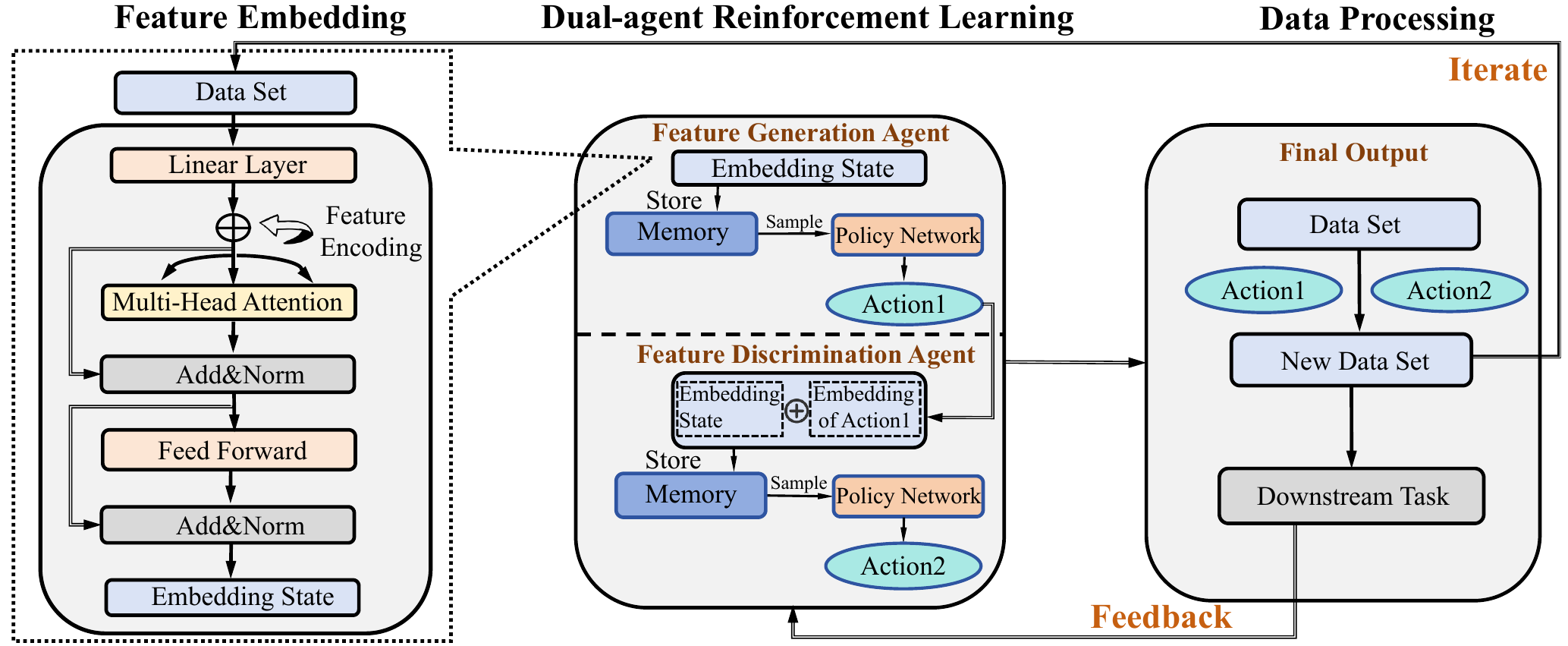} 
\caption{Overview of DARL. The dataset is transformed into feature embedding representations through a self-attention mechanism. Subsequently, a feature generation agent produces a sequence of operators, and a feature discrimination agent generates a discriminator sequence. These two sequences are combined with the original feature set to generate a new feature set. The updated feature set is then input into downstream tasks for evaluation, and the results are fed back to the two agents. This process iterates until the best feature set is discovered or the maximum number of iterations is reached.}
\label{fig_framework}
\end{figure*}

\section{Related Work}
The goal of automated feature generation is to identify the optimal feature set to enhance the performance of machine learning models \cite{chen2021techniques}. Feature generation methods include expansion-reduction methods and search-based methods. Expansion-reduction methods first expand the feature space and then perform feature selection to reduce redundancy. Search-based methods are a class of techniques used in automated feature generation that explore potential feature transformations and combinations using search methods to find the optimal feature set.

Expansion-reduction methods such as ExploreKit \cite{katz2016explorekit} include three stages in its workflow: candidate feature generation, candidate feature ranking, and candidate feature evaluation and selection. It executes all transformation functions on the complete dataset and selects the subset to be added based on the empirical performance of models trained with the candidate features. FEADIS \cite{dor2012strengthening}: it randomly selects original features and mathematical functions to generate new features. Autofeat \cite{horn2020autofeat}: it first generates a large number of nonlinear features and then selects a small subset of meaningful features from them to create a new dataset. It is evident that expansion-reduction methods struggle with the problem of feature space explosion. Currently, a more effective method is to use search methods to explore potential feature transformations in order to identify the most valuable features for predictive models.

In the field of search methods, GRFG \cite{wang2022group} is based on group-wise feature transformations. Through reinforcement learning, the system selects an operator for each group and then performs operations between the groups. TransGraph \cite{khurana2018feature} generates high-order features using transformation graphs and Q-learning algorithms. The Neural Feature Search (NFS) framework \cite{chen2019neural} assigns a recursive neural network controller \cite{jain2000recurrent} to each feature. Each feature generates a new feature independently. However, this method requires \textit{N} RNN controllers for \textit{N} features. Bigfeat \cite{eldeeb2022bigfeat} proposed a dynamic feature generation technique based on computing tree logic. These methods show promising results in automating feature generation and improving machine learning performance, but they generate excessive redundant features. While there are methods to improve relevance and reduce redundancy \cite{cond-weight,rel-var,shared-feat}, they have not been integrated into existing approaches. Additionally, the embedded states do not adequately represent the relationships between features. At the same time, they do not distinguish between discrete and continuous features. Our proposed method aims to address these limitations by integrating feature discrimination into the search process and enhancing the representation of feature relationships.

\section{The Proposed Method}

\subsection{Problem Settings}
Given a dataset $\mathcal{D} = \{ \mathcal{F}, y \}$, where $\mathcal{F}$ represents the original feature set and $y$ represents the target label set, the feature set $\mathcal{F} = \{D_1,\ldots,D_N,C_1,\ldots,C_M\}$, consists of $N$ discrete features $\{D_1,\ldots,D_N\}$ and $M$ continuous features $\{C_1,\ldots,C_M\}$. We aim to find the optimal feature set  $\mathcal{F}^*$ that maximizes: 
\begin{equation}
\mathcal{F}^* = argmax_\mathcal{F}(V_A(\mathcal{F},y)),
\end{equation}
where \textit{A} represents downstream machine learning tasks such as random forest, SVM, etc., and \textit{V} denotes evaluation metrics like \textit{F1-score}. Therefore, our objective is to identify a feature set that maximizes the performance metric of downstream tasks.

\subsection{Overall Framework}

In this section, we describe the overall framework idea of our feature generation algorithm. The framework adopts a hierarchical reinforcement learning strategy, which interacts with the environment through two Markov decision processes to achieve better overall effectiveness, as illustrated in Figure \ref{fig_framework}. 
Specifically, for a dataset $\mathcal{D}$, after feature embedding, it is input into the feature generation agent. The feature generation agent generates a new feature for each feature in the original feature set using predefined operations. The goal of this stage is to expand the feature set by creatively transforming existing features to increase the information content. Obviously, continuously generating a new feature for each feature would lead to a feature explosion, resulting in an excessive number of irrelevant features. Therefore, another agent is designed to eliminate these irrelevant features. Hence, we propose to design a feature discrimination agent. 
The feature discrimination agent aims to streamline the feature set by selecting the most useful features. 
Due to the hierarchical nature of the framework, the output of the feature generation agent serves as the input for the feature discrimination agent. The feature generation agent generates an operator (e.g., ``+", ``-") for each feature in the feature set $\mathcal{F}$, resulting in an operator sequence $\mathcal{T}_1 = \{p_1,\ldots,p_{M+N}\}$, while the feature discrimination agent generates a discriminator for each operator, resulting in a discriminator sequence $\mathcal{T}_2 = \{q_1,\ldots,q_{M+N}\}$. Combining the two action sequences together, $\mathcal{T} = \{p_1q_1,\ldots,p_{M+N}q_{M+N}\}$, generates a new action sequence. This new action sequence is then applied to the original feature set $\mathcal{F}$, resulting in a new, optimized feature set $\hat{\mathcal{F}}$. The new feature set $\hat{\mathcal{F}}$ is input into downstream tasks (e.g., random forest, SVM) for feature set evaluation and reward calculation. The evaluation results of downstream tasks are used as feedback rewards for the feature generation agent and the feature discrimination agent to guide them in obtaining better policy networks. To obtain higher-order feature representations, the original feature set $\mathcal{F}$ is updated to the newly generated feature set $\hat{\mathcal{F}}$, and the process is repeated for \textit{K} rounds until reaching the maximum number of iterations. Unlike traditional feature generation methods, our framework primarily aims to maximize cumulative rewards by treating feature discrimination as a Markov Decision Process (MDP). The framework automates the feature generation process and reduces the reliance on domain expert knowledge. It also considers computational efficiency and manages computational resources by limiting the number of iterations and optimizing algorithms.

\subsection{Feature Embedding}

Reinforcement learning, as a framework for solving the MDP, requires the description of the MDP state. The feature generation agent must take a state as the input, and this state is supposed to be constructed from the original feature set. Thus, the state representation from the original feature set to an embedding vector is mandatory. This section focuses on introducing the state of the feature generation agent. Previous methods primarily utilized autoencoders, graph convolutional autoencoders, etc., as the embedding representations of states \cite{xiao2022self}. These methods, with their shorter context lengths and limited ability to capture intricate dependencies, have not effectively learned the underlying relationships between features. Therefore, in this paper, we adopt the idea from the Transformer model \cite{vaswani2017attention} which has demonstrated excellent performance in embedding tabular data 
\cite{zhang2024tfwt,TabT}. It utilizes a multi-head self-attention mechanism to learn more effective embedding representations. Additionally, when processing tabular data, it is important to distinguish between discrete and continuous features.
The embedding representation of the state is shown in the feature embedding section of Figure \ref{fig_framework}. Similar to the architecture of a Transformer, the dataset is initially passed through a linear layer to transform the tabular data into a fixed dimension. Next, feature encoding is performed on the tabular data. Since tabular data does not convey positional information, feature encoding in this context only focuses on the distinctions between discrete and continuous features. The equation for feature encoding is shown as:

\begin{equation}
F_{\mathit{enc}} = \gamma \sin\left(\frac{p_f}{10^{\left(\frac{i}{{d}_\text{model}-1}\right)}}\right),
\end{equation}
where $p_f = -1$ for discrete features, $p_f = 1$ for continuous features, and $p_f = 0$ for labels. 
Considering that $sin(\omega t)$ is an odd function and has origin symmetry, it is chosen for feature encoding.
$\gamma$ is a coefficient used to amplify or diminish the influence of feature encoding. Adding positional encoding to feature embedding enables the distinction of different types of features.

After applying the multi-head self-attention mechanism as shown in Formula (\ref{attention}), followed by a residual connection with layer normalization, the output is fed into a feed-forward neural network. After that, another residual connection with layer normalization is applied to obtain the final embedding representation.

\begin{equation}\label{attention}
\text{Attention}(Q, K, V) = \text{softmax}\left(\frac{QK^T}{\sqrt{d_k}}\right) V.
\end{equation}

By utilizing the self-attention mechanism, the original tabular data can learn the distinct relationships between features. This type of state representation can lead to better results in reinforcement learning tasks.

\subsection{Dual-agent Reinforcement Learning}
We illustrate the proposed dual-agent reinforcement learning process in Figure \ref{fig_framework} in this section. The goal of the reinforcement learning agent is to find the optimal action by utilizing a policy network to maximize cumulative rewards. In a single training iteration, the agent receives the current state and generates actions through the policy network. The environment evaluates the actions to obtain rewards based on the machine learning algorithm \textit{A} and evaluation metric \textit{V}, and updates the policy network. Next, we introduce the states, actions, and rewards for the two agents.

\subsubsection{Feature generation agent.}
Feature generation agent learning system includes \textit{{i)} {state}} The state of the feature generation agent is defined as the embedded representation from the original tabular data through the self-attention mechanism, $s_1=emb(\mathcal{D})$. \textit{{ii) action}} The action space of the feature generation agent consists of operations on the original features. To distinguish discrete and continuous features, we utilize different action spaces. For continuous features, the action space includes operations such as ``absolute value", ``none", ``square", ``inverse", ``logarithm", ``square root", ``cube", ``addition", ``subtraction", ``multiplication" and ``division". For discrete features, the action space includes ``cross" \cite{luo2019autocross} and ``add". Feature cross means that if two original features are $f_1 = \{A, B\}$ and $f_2 = \{C, D\}$, then a new feature with four categorical values will be generated: $f_{new}=\{AC, AD, BC, BD\}$. By performing feature cross, the model can learn the combined effects of discrete features. To address the issue of cross-operation between discrete and continuous features, we convert continuous features into discrete features by binning with a decision tree \cite{ying2023self}. The binning process converts continuous features into discrete features so that they can interact with originally discrete features. This method allows for learning richer feature representations. \textit{{iii) reward}} The reward is calculated as the difference between the score achieved by the newly generated feature set $Score_{new}$ and the score of the original feature set $Score_{ori}$ in the downstream learning task. It can be calculated by:
\begin{equation}\label{e4}
r_1 = Score_{new} - Score_{ori}.
\end{equation}

\subsubsection{Feature discrimination agent.}
Feature discrimination agent learning system includes \textit{{i)} {state}} Considering the hierarchical relationship between the feature generation agent and the feature discrimination agent, in order to provide a better state representation for the feature discrimination agent, the action sequence generated by the feature generation agent is transformed into word embedding vectors, which are then concatenated with the state of the feature generation agent to form the state of the feature discrimination agent. This hierarchical representation enables the feature discrimination agent to consider the historical behavior of the feature generation agent, which helps capture the collaboration and dependency between agents. This hierarchical information enhances the contextual awareness of the feature discrimination agent's decision-making process, $s_2=emb(\mathcal{D}) \oplus  emb(\mathcal{T}_1)$. \textit{{ii) action}} The action space of the feature discrimination agent consists of ``delete", ``replace", and ``add". ``Delete" means that the newly generated feature should be removed. ``Replace" means that the newly generated feature is superior to the original feature and should substitute the original one. ``Add" means that the newly generated feature should be added to the original feature set. Through continuous evaluation and optimization, the feature discrimination agent will develop a long-term strategy to determine which features should be deleted, replaced, or added at each step, thereby creating the most valuable feature set. The feature discrimination agent can finely manage the feature set and offer a more refined state representation for the entire system through its decision-making process. The feature discrimination agent enhances the prediction accuracy and efficiency of the entire system. \textit{{iii) reward}} We design three reward functions $r_{del}$, $r_{rep}$, and $r_{add}$ corresponding to the three actions:

As Equation (\ref{del}) shows, to determine whether the newly generated feature $f_{new}$ should be deleted, the reward is formally represented as the difference between the mutual information of the original feature and the label $I(f_{ori}, y)$, and the mutual information of the newly generated feature and the label $I(f_{new}, y)$. A larger difference indicates that the original feature contains more information about the label $y$, therefore, the newly generated feature is meaningless and should be deleted.

\begin{equation}\label{del}
r_{del} = I(f_{ori},y) - I(f_{new},y).
\end{equation}

To determine whether the newly generated feature $f_{new}$ should replace the original feature $f_{ori}$, we design a reward function $r_{rep}$ using mutual information difference as Equation (\ref{rep}) shows. Opposite to $r_{del}$, a larger difference indicates that the new feature contains more information about the label $y$, therefore, the newly generated feature is meaningful and should replace the original feature.

\begin{equation}\label{rep}
r_{rep} = I(f_{new},y) - I(f_{ori},y).
\end{equation}

To determine whether the newly generated feature should be added, it is necessary to measure the redundancy between the original feature and the new one. We use mutual information between the original feature and the new feature $I(f_{ori},f_{new})$ to quantify their redundancy. A smaller mutual information value indicates a weaker correlation between the two features, suggesting lower redundancy. Therefore, the new feature should be added.

\begin{equation}\label{add}
r_{add} = I(f_{ori},f_{new}).
\end{equation}

The utility reward is defined as the difference between the score of the downstream learning task achieved using the newly generated feature set and the score achieved using the original feature set, as shown in Equation (\ref{imp}).

\begin{equation}\label{imp}
r_{imp} = Score_{new} - Score_{ori}.
\end{equation}

Combining the above equations, we obtain the reward for the feature discrimination agent as follows:
\begin{equation}\label{e9}
r_2 = \alpha*r_{del}+\beta*r_{rep}+\gamma*r_{imp} - \delta*r_{add},
\end{equation}
where $\alpha,\beta ,\gamma$ and $\delta$ are positive weights.

Mutual information is utilized as part of the reinforcement learning framework and works in conjunction with downstream task performance. This enables our method to more effectively balance short-term feature relevance and long-term task utility, overcoming the limitations of using mutual information alone for feature selection.

\subsubsection{Model Training.} 
We employ Deep Q-Networks (DQN) \cite{mnih2013playing} as the reinforcement learning model. DQN marries deep learning with Q-learning, using a neural network to approximate the Q-function, predicting the expected return of an action in a given state.
It balances exploration and exploitation using the $\epsilon$-greedy strategy, where the agent randomly selects actions with a probability of $\epsilon$ to explore the environment and with a probability of 1-$\epsilon$, selects the optimal action according to the current policy. Additionally, this method also employs the technique of experience replay, which involves maintaining a replay buffer to store tuples of data (state, action, reward and next state) sampled from the environment. During the training of the Q network, a random sample of data is drawn from the replay buffer for training purposes. Both agents need to minimize the Bellman formula of the action-value function and reduce the temporal difference error:

\begin{equation}\label{e10}
\mathcal{L} = Q(s_t,a_t) - (R(s_t,a_t) + \gamma max_{a_{t+1}}Q(s_{t+1},a_{t+1})).
\end{equation}

Here, $\gamma$ represents the discount factor, and \textit{Q} is the estimated \textit{Q} function by the deep neural network. As training progresses, the agent's policy gradually converges to the optimal policy $\pi^*$. This means that for a given state, the agent can choose the action that maximizes the expected cumulative return. It can be expressed as: 

\begin{equation}
\pi^* = argmax_a Q({s_t,a}).
\end{equation}

Through this training method, both agents learn optimal actions in their state spaces to maximize cumulative rewards. Information is exchanged between the two agents, and their decisions are guided by reward signals generated by the environment. These signals can be fed back to both agents to help them adjust their strategies. Through effective information exchange, the two agents form a more powerful and flexible system to better cope with complex and dynamic environments, thereby significantly improving the effectiveness of reinforcement learning.

\begin{table*}[ht]
\centering
\small
\setlength{\tabcolsep}{1mm}
\begin{tabular}{ccccccccccccc}
   
  \hline
  Datasets & Source & C/R & Samples/Features & Base & Random & PCA & DFS & Autofeat & Bigfeat & NFS & GRFG & DARL \\
  \hline
  PimaIndian & UCIrvine & C & 768/8 & 0.7566 & 0.7670 & 0.6444 & 0.7579 & 0.7566 & 0.7461 & 0.7806 & 0.7857 & \textbf{0.7904} \\
  German Credit & UCIrvine & C & 1001/24 & 0.7390 & 0.7620 & 0.5910 & 0.7610 & 0.7540 & 0.7370 & 0.7720 & 0.7740 & \textbf{0.7770} \\
  SPECTF & UCIrvine & C & 267/44 & 0.7751 & 0.8462 & 0.8051 & 0.7515 & 0.7856 & 0.8238 & 0.8500 & 0.8568 & \textbf{0.8688} \\
  Ionosphere & UCIrvine & C & 351/34 & 0.9233 & 0.9260 & 0.6893 & 0.9401 & 0.9381 & 0.9203 & 0.9516 & 0.9554 & \textbf{0.9601} \\
  Wine Quality Red & UCIrvine & C & 999/12 & 0.5395 & 0.5485 & 0.5135 & 0.5085 & 0.5275 & 0.5425 & 0.5736 & 0.5774 & \textbf{0.5856} \\
  Wine Quality White & UCIrvine & C & 4900/12 & 0.4976 & 0.5086 & 0.4710 & 0.4798 & 0.5021 & 0.4955 & 0.5117 & 0.5142 & \textbf{0.5202} \\
  Ilpd & OpenML & C & 583/10 & 0.6878 & 0.7342 & 0.6123 & 0.6929 & 0.6843 & 0.6980 & 0.7428 & 0.7387 & \textbf{0.7512} \\
  Svmguide3 & LibSVM & C & 1243/21 & 0.7989 & 0.8303 & 0.6678 & 0.8206 & 0.8230 & 0.8166 & 0.8339 & 0.8359 & \textbf{0.8391} \\
  Messidor Features & UCIrvine & C & 1150/19 & 0.6594 & 0.7411 & 0.5543 & 0.7524 & 0.7324 & 0.6498 & 0.7462 & \textbf{0.7482} & 0.7433 \\
  Lymphography & UCIrvine & C & 148/18 & 0.8170 & 0.8655 & 0.8577 & 0.8673 & 0.8444 & 0.8037 & 0.8717 & \textbf{0.8746} & 0.8715 \\
  \hline
  Airfoil & UCIrvine & R & 1503/5 & 0.5118 & 0.6106 & 0.4586 & 0.6074 & 0.5927 & 0.4980 & 0.6134 & 0.6197 & \textbf{0.6273} \\
  Housing Boston & UCIrvine & R & 506/13 & 0.4378 & 0.4597 & 0.2020 & 0.4760 & 0.4295 & 0.4203 & 0.4937 & 0.5012 & \textbf{0.5123} \\
  Openml\_586 & OpenML & R & 1000/25 & 0.6635 & 0.6321 & 0.3994 & 0.7187 & 0.7109 & 0.6635 & 0.7210 & 0.7310 & \textbf{0.7430} \\
  Openml\_607 & OpenML & R & 1000/50 & 0.6498 & 0.6367 & 0.2484 & 0.6814 & 0.6624 & 0.6363 & 0.6573 & 0.7005 & \textbf{0.7168} \\
  Openml\_618 & OpenML & R & 1000/50 & 0.6448 & 0.6194 & 0.2744 & 0.6848 & 0.6797 & 0.6351 & 0.6563 & 0.7071 & \textbf{0.7144} \\
  Openml\_592 & OpenML & R & 1000/25 & 0.6633 & 0.6578 & 0.2831 & 0.6939 & 0.6960 & 0.6633 & 0.6782 & 0.6926 & \textbf{0.7290} \\
  Openml\_584 & OpenML & R & 500/25 & 0.5826 & 0.5827 & 0.2153 & 0.5977 & 0.6356 & 0.5826 & 0.6020 & 0.6873 & \textbf{0.6929} \\
  Openml\_599 & OpenML & R & 1000/5 & 0.7199 & 0.7011 & 0.6344 & 0.7802 & 0.7233 & 0.7199 & 0.7819 & 0.7642 & \textbf{0.7937} \\
  \hline
  Bikeshare DC & Kaggle & R & 10886/11 & 0.9880 & 0.9920 & 0.9862 & 0.9993 & 0.9909 & 0.9995 & 0.9991 & 0.9994 & \textbf{0.9996} \\
  Ap-omentum-ovary & OpenML & C & 275/10936 & 0.7818 & 0.4550 & 0.5927 & 0.3787 & 0.4570 & 0.8072 & 0.8509 & 0.8691 & \textbf{0.8764} \\
  Adult Income & UCIrvine & C & 48842/14 & 0.8498 & 0.8494 & 0.5916 & 0.8483 & 0.8463 & 8.8295 & 0.8501 & 0.8505 & \textbf{0.8514} \\
  \hline
    
\end{tabular}
\caption{Overall performance comparison.}
\label{tab:over}
\end{table*}

\section{Experiments}

\subsection{Experimental Setup}
The experimental objectives include validating the improvement of our method compared to other feature generation methods, verifying the effectiveness of each part of the method through ablation experiments, demonstrating the stability of the algorithm on different downstream learning tasks, and comparing the runtime to validate algorithm time superiority. We conduct experiments on 21 datasets from UCI \cite{public2024}, Kaggle \cite{howard2024kaggle}, and OpenML \cite{public2024openml}, LibSVM \cite{lin2024libsvm}, comprising 12 classification tasks and 9 regression tasks. The evaluation metrics are as follows: for classification tasks, we used the \textit{F1-score} to evaluate recall and precision. For regression tasks, we use the 1-relative absolute error (1-RAE) to evaluate accuracy.

\subsection{Environmental Settings}
All experiments are conducted on
the Ubuntu operating system, Intel(R) Core(TM) i9-10900X CPU@ 3.70GHz, and V100, with the framework of Python 3.10.12 and PyTorch 1.13.1.

\subsection{Evaluation Metrics}
The equation for 1-RAE is as follows:
\begin{equation}
1-RAE = 1- \frac{\sum_{i=1}^n \lvert y_i - y_i^*\rvert}{\sum_{i=1}^n \lvert y_i - y_m\rvert},
\end{equation}
$y_i^*$ is the actual target value of the i-th observation, $y_i$ is the predicted target value of the i-th observation, and $y_m$ is the mean of all actual target values. 

\textit{F1-score} is the harmonic mean of Precision and Recall.
\begin{equation}
F_1 = 2 * \frac{Precision * Recall}{Precision + Recall}.
\end{equation}
Precision measures the proportion of positive predictions that are correctly labeled, defined as $Precision = TP / (TP + FP)$. \textit{TP} stands for true positives, while \textit{FP} stands for false positives. Recall measures the proportion of actual positive samples that are correctly identified by the model, given by: $Recall = TP / (TP + FN)$. \textit{FN} represents false negatives.

\subsection{Hyperparameter Settings}
The number of epochs is limited to 200. By using 6 exploration steps per epoch, we further control the number of features generated. We adopt random forest as the downstream machine learning model and performed 5-fold stratified cross-validation in all experiments, instead of a simple 70\%-30\% split. We used the Adam \cite{adam} optimizer with a learning rate of 0.0001 to optimize DQN, and set the memory limit of experience replay to 24, and the DQN batch size to 8. The model incorporated 8 attention heads, with a word embedding vector dimension of 8 and a model hidden layer dimension of 128. The discrimination agent's reward weights $\alpha$, $\beta$, $\gamma$, and $\delta$ are set to 0.1, 0.1, 1, and 0.01.

\subsection{Baseline Methods}
We compare our method with 8 widely used feature generation methods, as well as random generation and feature dimension reduction methods: (1) Base: using the original dataset without feature generation. (2) Random: randomly generating features for each feature. (3) DFS \cite{kanter2015deep}: an expansion-reduction method that first expands and then selects feature, automatically generated features for the dataset. (4) PCA \cite{c:24}: a feature reduction method that compresses the original feature set. (5) Autofeat \cite{horn2020autofeat} is an expansion-reduction algorithm that generates nonlinear features and selects a subset of relevant features to form a new dataset. (6) Bigfeat \cite{eldeeb2022bigfeat}: proposes a dynamic feature generation technique based on computational tree logic. (7) NFS \cite{chen2019neural}: uses the RNN as a controller for each original feature, which outputs actions to generate new features. (8) GRFG \cite{wang2022group}:  iteratively generates new features and reconstructs an interpretable feature space through group-group interactions. We conduct experiments on the open-source code provided by these methods.

\subsection{Overall Comparison}
Table \ref{tab:over} shows the comparison of our method
with eight baseline models on the 21 datasets in terms of \textit{F1-score} or 1-RAE. Our method outperformed the others on most datasets. Additionally, to validate the performance of the proposed method on large datasets where both the feature and sample sizes exceed 10,000, we conducted experiments on three large datasets. The results confirmed the effectiveness of our proposed method. It can be observed that our dual-agent feature generation method outperforms other feature generation methods by maximizing cumulative rewards through reinforcement learning.

\begin{table*}[t]
\centering
\small 
\setlength{\tabcolsep}{1mm}
\begin{tabular}{cccccccccc}
  \hline
   Feature Types and Proportions& Openml\_584 & Openml\_599 & Housing Boston & Airfoil & Ilpd & Ionosphere & German Credit & PimaIndian \\
  \hline
  Low-order features & 33 & 11 & 9 & 11 & 11 & 0&1&2 \\
  High-order features & 147 & 26 & 47 & 175 & 10 & 13 &38&4 \\
  Proportion of high-order features & 81.7\% & 70.3\% & 83.9\% & 94.1\% & 47.6\% &100.0\%&97.4\%&66.7\% \\

  \hline
    
\end{tabular}
\caption{Proportion of high-order features generated by different datasets.}
\label{tab:pro}
\end{table*}

\subsection{Ablation Study}

This experiment aims to verify whether each component of our method indeed has a positive impact on the final results. Therefore, we have developed three variants: (1) The feature discrimination agent was replaced with mutual information feature selection. This change was made to verify whether constraining feature quantity through the feature discrimination agent is more effective than using mutual information.
This variant is referred to as ``DARL-k". (2) The state of the feature generation agent does not utilize self-attention encoding. This is to verify if self-attention encoding is more effective. This variant is referred to as ``DARL-t". (3) For discrete features, no cross-operation and feature encoding are performed to examine the necessity of distinguishing discrete features from continuous features. This variant is referred to as ``DARL-c". We validate these variants on four datasets. Tabel \ref{tab:ablation} shows that the best performance is achieved when all components are present. This validates the effectiveness of our method: dual-agent reinforcement learning is more effective than traditional feature selection methods, as reinforcement learning can maximize cumulative rewards. Self-attention encoding is more effective than not using it, as it considers deep connections between features. Distinguishing between feature types is more effective than not distinguishing them, as there are inherent differences between discrete and continuous features. It is noteworthy that not distinguishing between discrete and continuous features yields similar results on regression datasets. This is because regression task datasets often lack discrete features, only containing continuous features, making the differentiation between discrete and continuous features irrelevant.
\begin{table}[H]
\centering
\small
\setlength{\tabcolsep}{1mm}
\begin{tabular}{ccccc}
 \hline
  Datasets & DARL & DARL-k & DARL-t & DARL-c \\
  \hline
  Airfoil & \textbf{0.6273} & 0.5244 & 0.6241 & \textbf{0.6273} \\
  German Credit & \textbf{0.7770} & 0.7750 & 0.7650 & 0.7680 \\
  Housing Boston & \textbf{0.5123} & 0.4811 & 0.4982 & \textbf{0.5123} \\
  PimaIndian & \textbf{0.7904} & 0.7891 & 0.7892 & 0.7800 \\
  
  \hline
    
\end{tabular}
\caption{Comparison of different DARL variants in terms of \textit{F1-score} or 1-RAE.}
\label{tab:ablation}
\end{table}

\subsection{Robustness Analysis}

\begin{table}[ht]\label{overall}
\centering
\small
\setlength{\tabcolsep}{1mm}
\begin{tabular}{c|cc|cc|cc}
   \hline
   \multicolumn{1}{c|}{\centering Datasets} & \multicolumn{2}{c|}{XGB} & \multicolumn{2}{c|}{SVM} & \multicolumn{2}{c}{Logistic} \\
   
   & Base & DARL  & Base & DARL & Base & DARL \\
   \hline
   Airfoil & 0.4692 & \textbf{0.6301} & 0.0340 & \textbf{0.0430} &  0.2792 & \textbf{0.3406}\\
   SPECTF & 0.8015 & \textbf{0.8313} & 0.7901 & \textbf{0.8050}  & 0.8014 & \textbf{0.8238} \\
   Ilpd & 0.6827 & \textbf{0.7067} & 0.7086 & \textbf{0.7153}  & 0.7187 & \textbf{0.7221} \\
   PimaIndian & 0.7513 & \textbf{0.7566} & 0.7693 & \textbf{0.7826} & 0.7696 & \textbf{0.7800}\\
   \hline
\end{tabular}

\caption{Comparison of different machine learning models in terms of \textit{F1-score} or 1-RAE.}
\label{tab:rob}
\end{table}

This experiment aims to compare the performance improvement achieved by using various machine learning algorithms to process the new dataset against the original dataset. This will help verify the robustness of the method. We employed XGBoost (XGB), Support Vector Machine (SVM), and Logistic Regression (Log) on the new dataset to validate if the newly generated feature set still enhances performance compared to the original dataset. As Table \ref{tab:rob} shows, we conducted experiments on 4 datasets and observed that the scores of the new datasets were consistently higher than those of the original datasets, indicating the high robustness of our method.

\subsection{Efficiency Improvement}

\begin{figure}[H]
    \centering
    \begin{subfigure}[b]{0.233\textwidth}
        \centering
        \includegraphics[width=\textwidth]{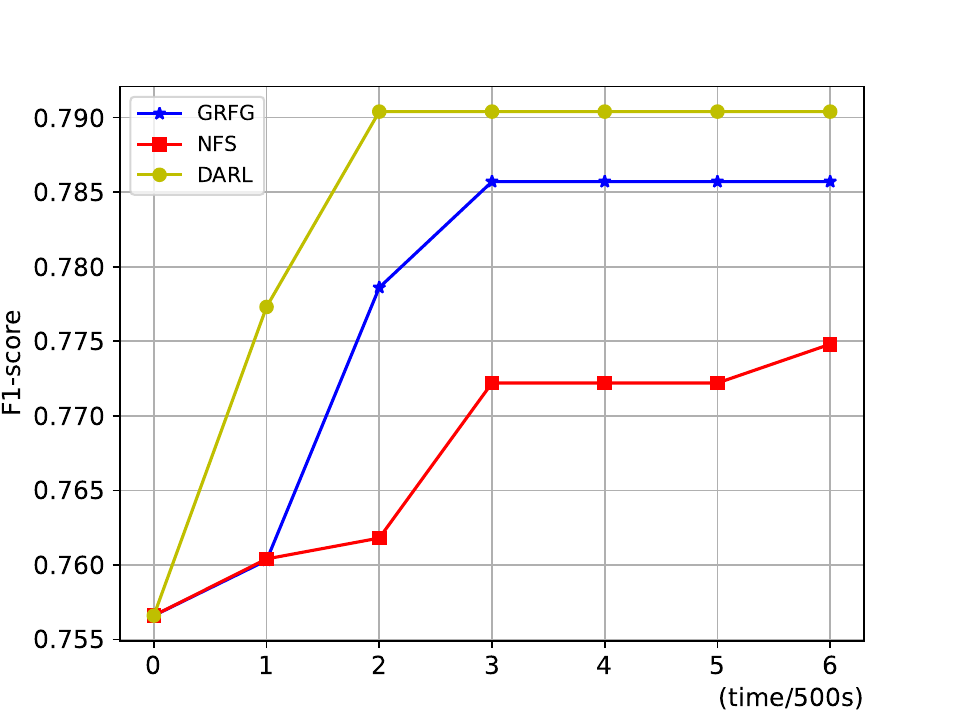}
        \caption{PimaIndian}
        \label{fig:image1}
    \end{subfigure}
    \begin{subfigure}[b]{0.233\textwidth}
        \centering
        \includegraphics[width=\textwidth]{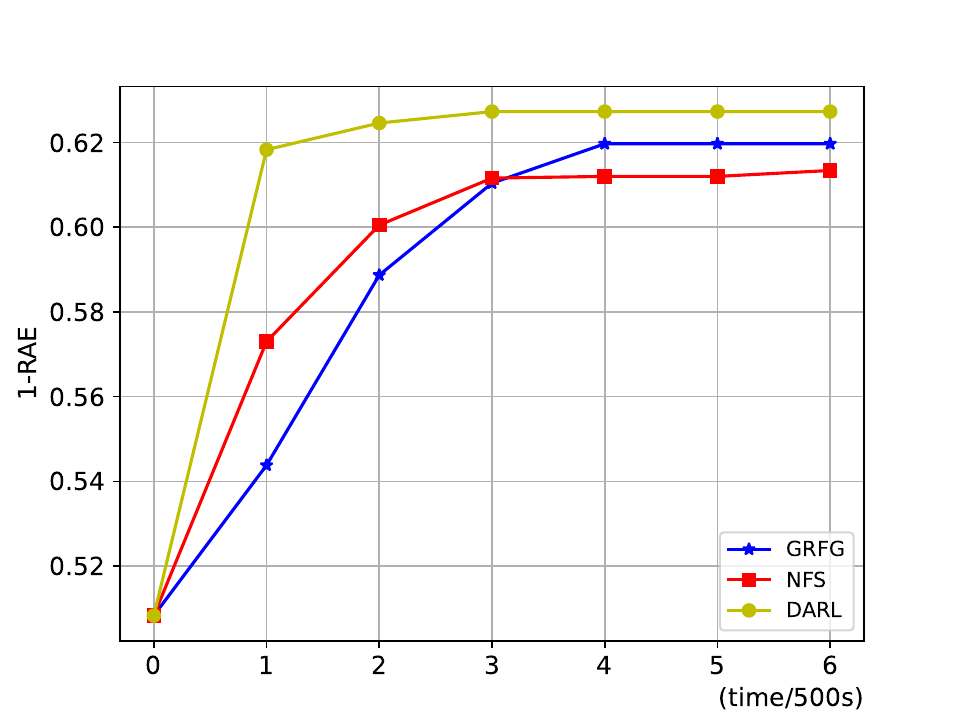}
        \caption{Airfoil}
        \label{fig:image2}
    \end{subfigure}

    \caption{Time comparison of different algorithms.}
    \label{fig:four_images}
    
\end{figure}

This experiment aims to verify the optimality of our method in terms of efficiency. By comparing the running time of our method with NFS and GRFG on two datasets, Figure \ref{fig:four_images} shows that our method consumes the least amount of time and achieves the best scores.

\subsection{Feature Order Analysis}
High-order features can capture more complex patterns and relationships in the data. The experiments in Table \ref{tab:pro} aim to determine whether the newly generated dataset contains high-order features. We define features with an order greater than or equal to 2 as high-order features. Experiments conducted on 8 datasets show that on most of them, more than 70\% of the features are high-order features.

\section{Conclusion}
In this paper, we propose a novel dual-agent reinforcement learning (DARL) method for feature generation. This method aims to capture complex relationships between features to enhance the performance of machine learning tasks.  In DARL, two agents are designed for the reinforcement learning framework, where the first agent generates new features, and the second agent determines whether the generated features are worth preserving. We propose to use a self-attention mechanism to enhance state representation, effectively capturing the complex relationships between features. We propose to distinguish between discrete and continuous feature interactions to generate more interpretable features. Extensive experiments have demonstrated that DARL exhibits significant effectiveness in feature generation when compared to other baseline methods. Pseudocode of the DARL, experimental settings, comparison of different downstream task and convergence analysis are presented in Appendix.

\section{Acknowledgments}
This work was supported by the Science Foundation of Jilin Province of China under Grant YDZJ202501ZYTS286, and in part by Changchun Science and Technology Bureau Project under Grant 23YQ05.

\bibliographystyle{named}
\bibliography{ijcai25}

\newpage
\appendix

\section{Pseudocode of the Algorithm}
Algorithm \ref{al} in pseudocode processes dataset $\mathcal{D}$ via feature generation and discrimination agents, yielding action sequences. These are merged with $\mathcal{D}$ to create a new dataset. Rewards for each agent are computed using Equation (4) and (9). The policy network is updated until maximum iterations are met.
\begin{algorithm}[H]
\caption{DARL algorithm}
\textbf{Input}: Original table data $\mathcal{D}$, downstream machine learning task \textit{A}, evaluation metric \textit{V}.\\
\textbf{Parameter}: Number of reinforcement learning epochs \textit{E}, exploration steps per epoch \textit{K}.\\
\textbf{Output}: New dataset.
\begin{algorithmic}[1] 
\STATE Let $e=0$.
\WHILE{epoch $e  < E$}
\STATE Let $k=0$.
\WHILE{ step $k  < K$}
\STATE Sequence of operators $\mathcal{T}_1$ generated by the feature generation agent.
\STATE Discriminator sequence $\mathcal{T}_2$ generated by the feature discrimination agent.
\STATE Generate a new dataset $\mathcal{D}$ based on $\mathcal{T}_1$ and $\mathcal{T}_2$.
\STATE Calculate the reward for the feature generation agent according to Equation (4).
\STATE Calculate the reward for the feature discrimination agent according to Equation (9).
\STATE Update the policy network according to Equation (10).
\ENDWHILE
\ENDWHILE
\STATE \textbf{return} New dataset.
\end{algorithmic}
\label{al}
\end{algorithm}

\section{Comparison of Different Downstream Tasks}
This experiment evaluates the degree of algorithmic improvement by substituting different downstream machine learning (ML) algorithms, including XGBoost (XGB), LightGBM, and Logistic Regression (Log). It can be observed that these algorithms show equivalent degrees of improvement compared to the previous random forest algorithm. Tabel \ref{tab:down} demonstrates that our algorithm can enhance performance universally across different types of downstream machine learning algorithms, regardless of their specific types.

\begin{table}[H]  
\centering  
\small 
\setlength{\tabcolsep}{1mm}
\begin{tabular}{c|ccccccccc}
    \hline 
   ML Algorithms& Base & Random & PCA & DFS & Autofeat  \\ 
 \hline 
XGB & 0.7501& 0.7720& 0.6548& 0.7630& 0.7436 \\  

LightGBM & 0.7605& 0.7813& 0.6601 & 0.7592 &0.7657 \\   

Log  & 0.7696 & 0.7617 & 0.6562 & 0.7604 & 0.7384  \\  
\hline 
ML Algorithms& Bigfeat & NFS & GRFG & DARL\\
\hline 
XGB & 0.7552&  0.7827 & 0.7874& \textbf{0.7970}\\
LightGBM&0.7696&  0.7813 & 0.7897& \textbf{0.8073}\\
Log & 0.7812 &  0.7748 & 0.7839& \textbf{0.7995} \\
\hline 
\end{tabular}  
\caption{The performance of different downstream machine learning tasks, when using various feature generation algorithms, on the PimaIndian dataset (measured by F1-score or 1-RAE).}
\label{tab:down}
\end{table}

\section{Convergence Analysis}
\begin{figure}[H]
\centering
\includegraphics[width=0.4\textwidth]{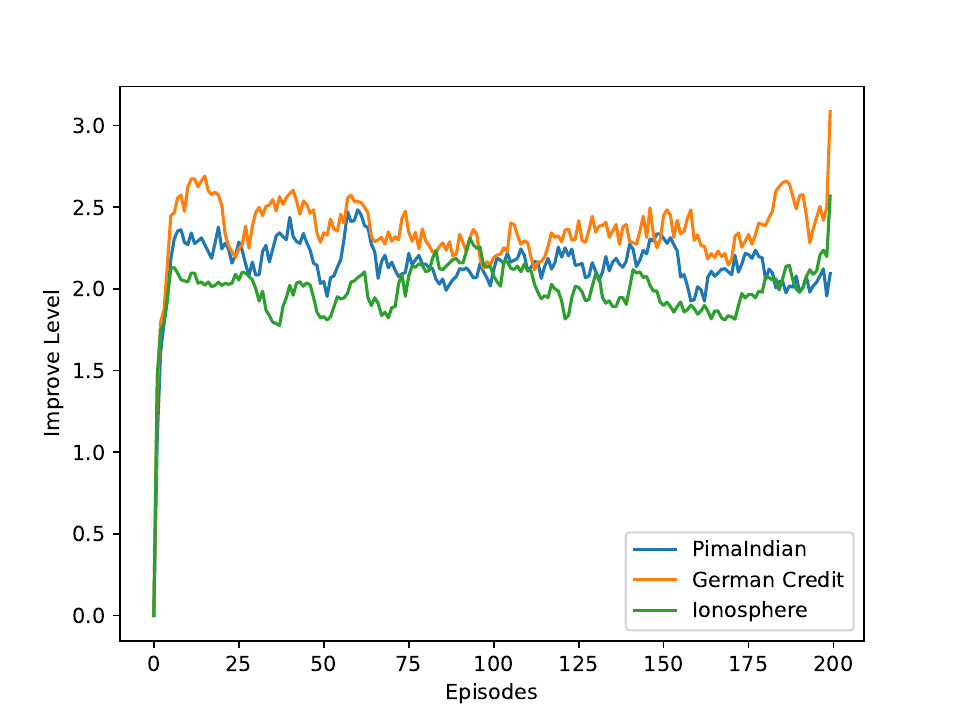} 
\caption{The convergence of the algorithm.}
\label{fig3}
\end{figure}

This experiment aims to verify the convergence of reinforcement learning, specifically whether reinforcement learning can steadily converge while enhancing performance. Figure \ref{fig3} depicts the results of experiments conducted on three datasets, showing that good results can be achieved after approximately 50 epochs. This experiment demonstrates that our method effectively achieves convergence.

\end{document}